\documentclass[letterpaper, 10 pt, conference]{ieeeconf}  
\IEEEoverridecommandlockouts                            
\usepackage{graphics} 
\usepackage{epsfig} 
\usepackage{mathptmx} 
\usepackage{times} 
\usepackage{amsmath} 
\usepackage{amssymb}  
\usepackage{graphicx}
\usepackage{multirow}
\usepackage{booktabs}
\usepackage{array}
\usepackage{stfloats}
\usepackage{url}
\usepackage{verbatim}
\usepackage{cite}
\usepackage{algorithm}
\usepackage{caption}
\usepackage{subcaption}

\title{\LARGE \bf
Parsimonious Dataset Construction for Laparoscopic Cholecystectomy Structure Segmentation
}
\author{Yuning Zhou$^{1}$,
Henry Badgery$^{3}$,
Matthew Read$^{3,4}$, 
James Bailey$^{2}$, 
Catherine Davey$^{1}$,
\thanks{$^{1}$Department of Biomedical Engineering, The University of Melbourne, Melbourne, VIC, Australia}
\thanks{$^{2}$School of Computing and Information Systems, The University of Melbourne, Melbourne, VIC, Australia}
\thanks{$^{3}$Department of HPB/UGI Surgery, St Vincent’s Hospital, Melbourne, Australia}        
\thanks{$^{4}$Department of Surgery, The University of Melbourne, St Vincent’s Hospital, Melbourne, Australia}
\thanks{email: yunizhou@student.unimelb.edu.au}}      

\begin{document}

\maketitle

\thispagestyle{empty}
\pagestyle{empty}

\begin{abstract}
Labeling has always been expensive in the medical context, which has hindered related deep learning application. Our work introduces active learning in surgical video frame selection to construct a high-quality, affordable Laparoscopic Cholecystectomy dataset for semantic segmentation. Active learning allows the Deep Neural Networks (DNNs) learning pipeline to include the dataset construction workflow, which means DNNs trained by existing dataset will identify the most informative data from the newly collected data. At the same time, DNNs' performance and generalization ability improve over time when the newly selected and annotated data are included in the training data. We assessed different data informativeness measurements and found the deep features distances select the most informative data in this task. Our experiments show that with half of the data selected by active learning, the DNNs achieve almost the same performance with 0.4349 mean Intersection over Union (mIoU) compared to the same DNNs trained on the full dataset (0.4374 mIoU) on the critical anatomies and surgical instruments. 
\newline

\indent \textit{Clinical relevance}— This paper investigates an efficient and scalable dataset construction method for laparoscopic cholecystectomy structure recognition. 
It will provide a premise for deep learning applications in clinical practice.
\end{abstract}

\section{Introduction}

Laparoscopic cholecystectomy (LC) is a commonly performed minimally invasive surgery to remove the diseased gallbladder, where surgeons operate under the guidance of the laparoscopic video transmitted to the monitor. 
To mitigate the primary complication, bile duct injury (BDI), introduced by the structure misidentification, deep learning has been used recently to improve critical anatomical structures and instrument recognition by highlighting them in real-time surgical video frames, known as semantic segmentation. 
A big challenge in existing studies and thus limited clinical usage is the lack of large-scale datasets due to annotation difficulties. 
Current public LC segmentation datasets are all constructed using naive frame selection methods on video recordings, including continuous selection \cite{cholecseg8k}, random selection \cite{mascagni2022artificial,madani2022artificial}, and thresholded selection based on pixel-wise differences \cite{tokuyasu2021development}. 
These methods are either too naive without considering the visual differences, or require cumbersome human calibration.
They failed to consider the frames relations from different videos, therefore select massive duplicated frames \cite{cholecseg8k} and reducing diversities and causing information loss.
Moreover, following the normal dataset construction workflow, existing methods collect all surgical recordings first and process them with the same selection criterion, which consider neither the dataset construction efficiency nor class balance.

\begin{figure}[H]
    \centering
    \includegraphics[scale=0.32,trim={0.0cm 0.0cm 0.0cm 0.0cm}]{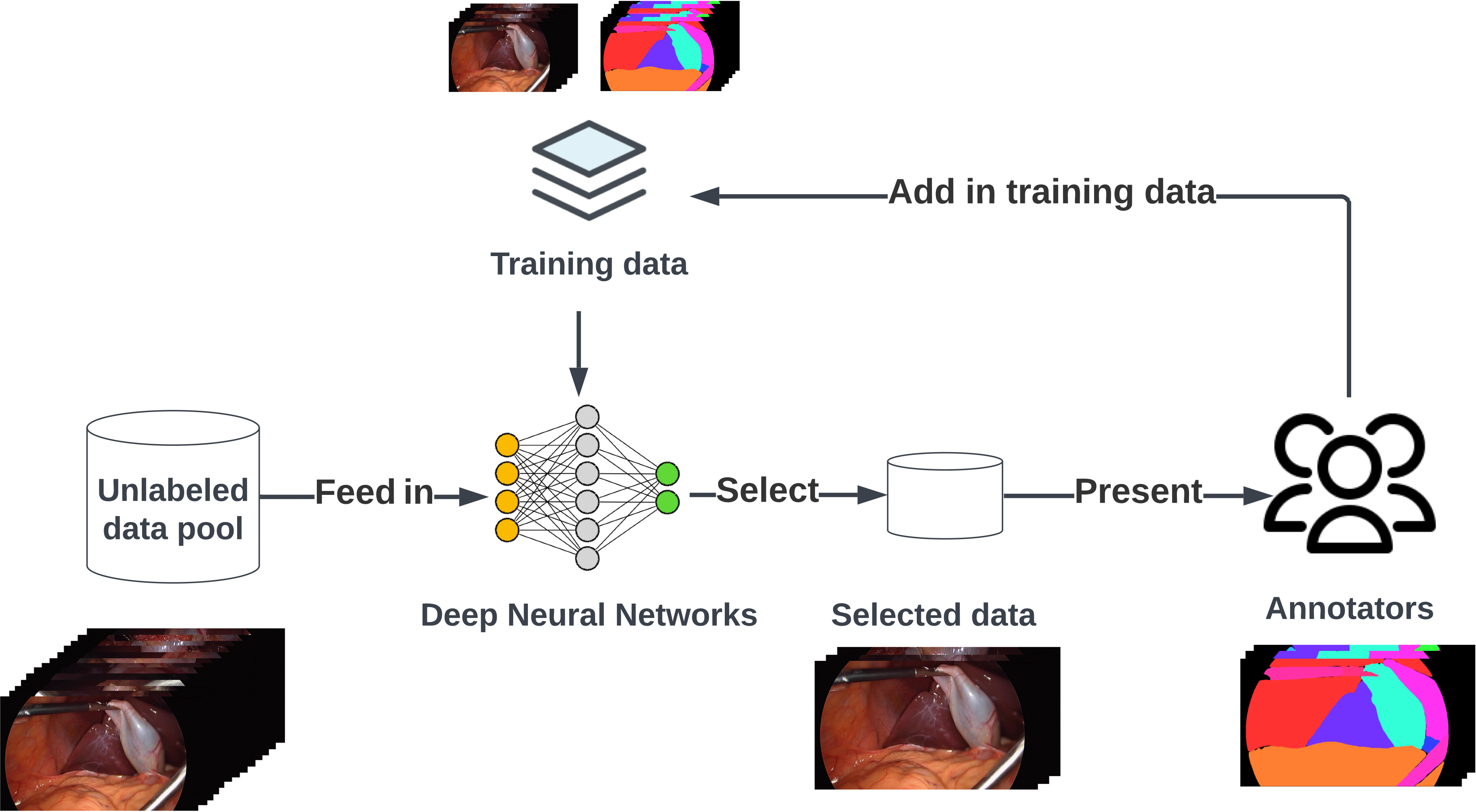}
    \caption{Illustration of active learning process}
    \label{fig:plot1}
\end{figure}

Recently, active learning has been demonstrated effectively minimizing the dataset size while maximizing the model's performance in medical classification datasets \cite{smailagic2018medal}.
Active learning assumes that the model and the training dataset will evolve over time as illustrated in Figure \ref{fig:plot1}, where in each iteration the model will be about to sample informative data from the unlabeled data pool, which will then be present to the annotators for labelling and fed back to the model training \cite{cohn1996active}.
In our paper, we adopt active learning for the first time as a parsimonious frame selection strategy in LC segmentation.
Specifically, we apply active learning to consider newly introduced subject at each round, where the deep learning model (DNN) selects the unlabeled and informative data subset by informativeness measurement strategies on the subject's video frames.
The selected frames will then be annotated and included in the training dataset to retrain the model at each round.
We access different informativeness measures from both the uncertainty aspect and diversity aspect.
Trained on the data selected by active learning, the model shows better structure identification ability than random selection.
Moreover, compared with random selection, using active learning could achieve the same level of performance by just half of the data.  
The results illustrate the possibility of constructing quality deep learning datasets within the affordable annotation budget and efficient pipeline from massive video data, which can fertilize more deep learning applications in surgery assistance.

For the rest of the paper, Section \ref{sec:methods} will introduce the informativeness measurement strategies and the experiment settings. Results will be demonstrated in Section \ref{sec:results}, with detailed analysis. In the last section \ref{sec:conclusion}, we will summarize our work and its possible contribution to this interdisciplinary challenge.

\section{Methods}

\label{sec:methods}
In each active learning selection round, the unlabeled frames will be fed into the model, where some informativeness measurement strategies, also known as acquisition functions, will be applied for ranking and sampling \cite{cohn1996active}.
Intuitively, an unlabeled frame is considered significant from two aspects: when the model is unconfident in its prediction, known as uncertainty \cite{li2013adaptive,joshi2009multi}; and the frame itself is different from the existing dataset distribution, known as diversity \cite{smailagic2018medal,sener2017active}.
We access these two types of informativeness measurement strategies and their performance in the context of LC surgery.
Specifically, we use prediction entropy as the uncertainty measurement, and feature distance as diversity measurement.

\subsection{Prediction entropy}
In semantic segmentation, for an input image with height $H$ and width $W$, the network output will be in the shape $K \times H \times W$, where $K$ refers the total number of classes defined in the task. 
That is, for each pixel, it will be predicted as each of the $K$ classes with some probability $p$, where $\sum_{i=1}^{K} p_{i} = 1$.
We use the prediction's entropy as the uncertainty measurement to evaluate the model's confidence in predicting a particular unlabeled frame. 
For a pixel, the entropy is defined as:
\begin{equation}
H = -\sum_{i=1}^Kp_{i}\log(p_{i}).
\end{equation}
If the predicted probability of a pixel distributes more evenly on all the classes, the entropy will be higher.
For a whole image with $H \times W$ pixels, we calculate the average entropy of all pixels as the uncertainty measures.
After calculating the mean entropy of every unlabeled frame in a new video, we rank and select frames in decreasing mean entropy order.
To avoid selecting frames with similar visual representation, we further divide them into batches, and select a fix number of frames randomly from each batch \cite{kirsch2019batchbald}.

\subsection{Deep feature representation distance}
Unlike input frame pixel difference calculation, we evaluate the deep feature representation extracted by the trained DNNs at each round to sample frames with broader distributional diversities as they contain more information \cite{smailagic2018medal}.
Ma et al.
\cite{Characterizing:LID} shows deep features of the training data will lies in concentrated surroundings in the high dimensional manifold.
Correspondingly, for an unlabeled frame, the shorter feature distance from the existing data indicates similar feature representation. Conversely, the longer feature distance from others suggest that the feature representation do not lie on the manifold. 
It can be assumed that such frame is more distinct from the existing dataset, therefore more critical to include to expand the data diversities.

We accessed the \emph{Euclidean} (eq.\ref{ed}) and \emph{Cosine} (eq.\ref{cs}) distance respectively as diversity measurement for the feature representations.
Specifically, we pass each unlabeled data into the DNNs and extract the deep feature representation from the intermediate layer.
For a deep feature representation in $d$ dimensional sapce, we indicate it as $F \in \ \mathbb{R}^{d}$.
An unlabeled frame's deep feature $F_q$ is treated as a query. We calculate its pairwise distance with every existing labeled data's feature $F_r$.

\emph{Euclidean distance} measures the distance between two feature representations by considering the difference in every dimension:
\begin{equation}\label{ed}
Euclidean(F_q,F_r) = \sqrt{\sum_{i}^{d}(F_{qi}-F_{ri})^2}.
\end{equation}

\emph{Cosine distance} regards the two feature representations as two vectors and measures the similarity by calculating the inner product:

\begin{equation}\label{cs}
Cosine(F_q,F_r) = \frac{F_q \cdot F_r}{|F_q||F_r|}.
\end{equation}

After calculating the feature distance of each query data regarding all the reference data, we average the distances as the final distance score of the corresponding unlabeled frame.

Additionally, we not only favor the more dissimilar unlabeled frames from the existing dataset but also want to increase diversities by not selecting the similar frames from one video.
Therefore, we also consider the feature distance between unlabeled frames.
To distinguish the two feature distances, we refer to the one calculated between the unlabeled frame and the training data as the \emph{inter-distance} and the one calculated within the video frames as the \emph{intra-distance}.
We combine the two distances by summarizing the normalized \emph{inter-} and \emph{intra-distance}, and select frames based on the decreasing order of the sum.

\subsection{Dataset preperation}
In our study, five subjects' LC operation video records with ethical approval are selected, de-identified, trimmed down to the duration with only the risky duration for the BDI occurrence. 
They are then broken down into separate frames (in 32 frame-per-second), where blurry frames are removed. 
A pixel-wise Euclidean Distance threshold is applied further to avoid excessive data while retaining the frame variety and consistency, ended up with in total of 607 frames. For the five subjects, we ruled a random one out as a hold-out test subject. Finally, our dataset contains 440 frames in the training and validation dataset, and 167 for the hold-out test set.  

To simulate the active learning data selection from unlabeled data pool, we use training data from annotated set as unlabeled pool. In each active learning iteration we present the data from unlabeled pool for selections.
Selected frames will be removed from the pool in the next iterations. This simulates these frames are annotated by experts.  

\subsection{Experiment settings}
We use the data from the first video as the initial round, and randomly split the frames in 80\% and 20\% as the training and validation data to train an initial model, DeepLabV3+ \cite{chen2018encoder} with ResNet101 \cite{he2016deep} as the feature extractor. 
Then, in each frame selection round when we consider to include a new video in the dataset, we access different informativeness frame selection strategies and compare them with random selection as the baseline selection strategy. 
We aim to include 50 new frames from each newly considered video.
The selected data are then paired with their labels and included in the training set, augmented and feed back in to the model training.
The model performances are accessed after each selection round in a ruled-out test video to evaluate the model's generalization ability on unseen subject.
Intersection out of Union (IoU), the standard region-based evaluation metric, is adopted to evaluate the model performance by measuring the overlapping area between the predictions and the annotations.

\section{Results and Analysis}
\label{sec:results}
\begin{figure}[H]
    \centering
    \includegraphics[scale=0.3,trim={0.0cm 0.0cm 0.0cm 0.0cm}]{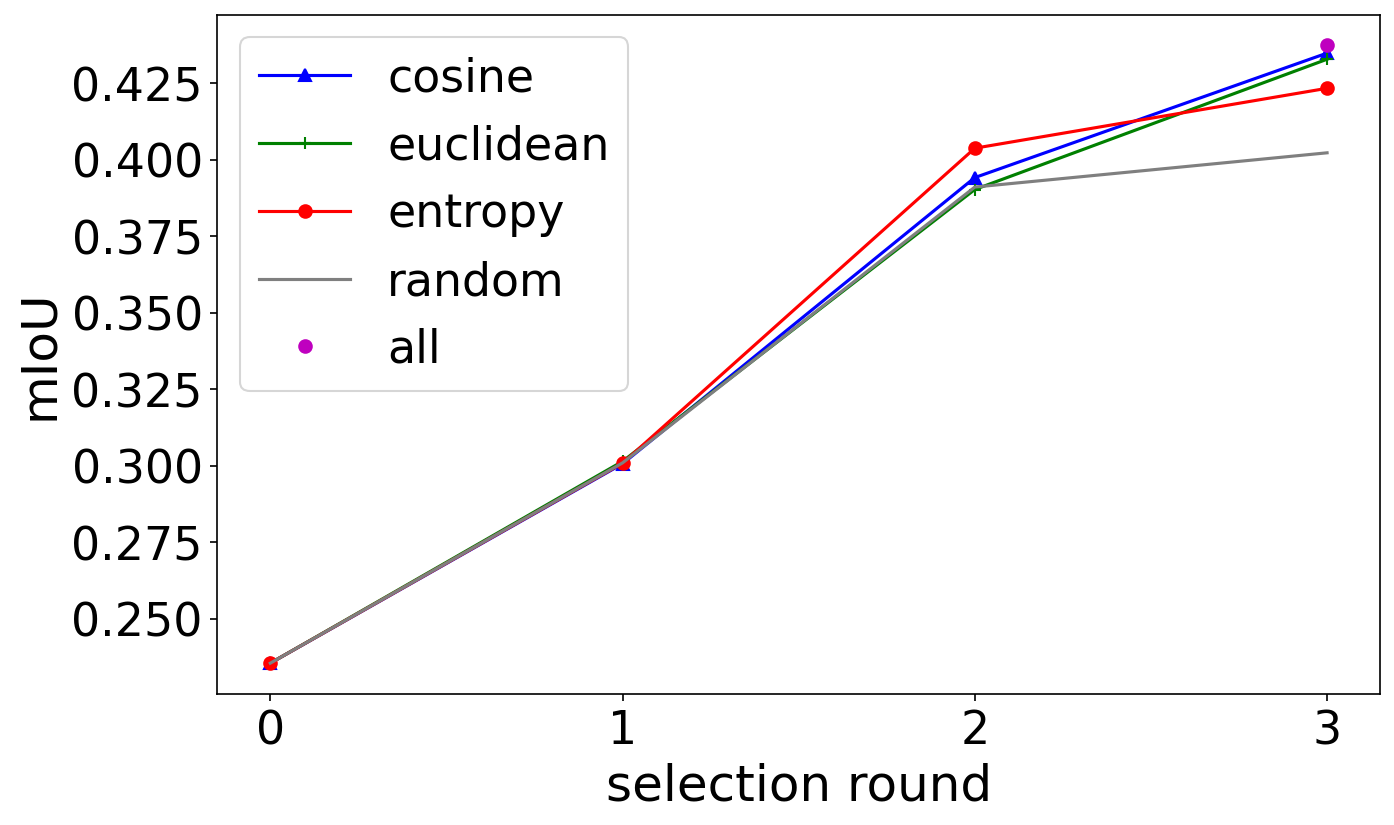}
    \caption{Comparison of DNNs performance trained under 4 frame selection methods for 3 active learning iterations}
    \label{fig:plot2}
\end{figure}

Due the the preliminary dataset limitation, the active learning frame selection only run for 3 rounds (considering one video for model initialization and one for testing).
The performance comparison of DeepLabV3+ trained under 3 active learning strategies with random selection as the baseline is shown in Figure \ref{fig:plot2}. 
The x-axis refers to the active learning selection round from the initial training set (\emph{round 0}). 
The y-axis represents the mean IoU of the 25 defined classes.
In the legend, \emph{random} indicates random selection.
\emph{entropy}, \emph{euclidean} and \emph{cosine} represent the three selection strategies.
\emph{all} indicates the model performance trained on full training set.

The figure indicates that after the initial stage, two feature distance measurement strategies both outperform the baseline by an increasingly significant margin. 
The entropy also performs better than the baseline while both of these two strategies show a converging trend.
It indicates that for the strategies we accessed, the diversity-based strategies can select more informative frames than the uncertainty-based frame selection strategy in LC surgical segmentation.

\begin{table}[!h]
\caption{Class-wise performance comparison for 6 critical structures and instruments (out of 25). \emph{All data} refers to the result for the full training set. Best results are in \textbf{bold} for each round.}
\label{tab:my-table}
\resizebox{\columnwidth}{!}{%
\begin{tabular}{@{}lccccccc@{}}
\toprule
\textbf{\begin{tabular}[c]{@{}l@{}}Round \&\\ strategies\end{tabular}} & \multicolumn{1}{l}{\textbf{\begin{tabular}[c]{@{}l@{}}Total frame \\ numbers\end{tabular}}} & \textbf{\begin{tabular}[c]{@{}c@{}}Cystic \\ duct\end{tabular}} & \textbf{Gallbladder} & \textbf{\begin{tabular}[c]{@{}c@{}}Segment \\ IV\end{tabular}} & \textbf{\begin{tabular}[c]{@{}c@{}}Diathermy \\ hook tip\end{tabular}} & \textbf{\begin{tabular}[c]{@{}c@{}}Johann \\ grasper \\ tip\end{tabular}} & \textbf{\begin{tabular}[c]{@{}c@{}}Scissors \\ tip\end{tabular}} \\ \midrule
\textbf{Init}                                                          & 72                                                                                    & 0.0123                                                          & 0.5669               & 0.2526                                                         & 0.6146                                                                 & 0.3576                                                                    & 0.0                                                              \\ \midrule
\textbf{R1 random}                                                     & \multirow{4}{*}{122}                                                                  & 0.0915                                                          & \textbf{0.6301}      & 0.4689                                                         & 0.7906                                                                 & 0.4380                                                                    & 0.0                                                              \\
\textbf{R1 entropy}                                                             &                                                                                       & 0.0751                                                          & 0.6071               & 0.4349                                                         & 0.7787                                                                 & 0.4262                                                                    & 0.0                                                              \\
\textbf{R1 euclidean}                                                           &                                                                                       & 0.0990                                                          & 0.5996               & \textbf{0.4707}                                                & 0.7852                                                                 & 0.4420                                                                    & 0.0                                                              \\
\textbf{R1 cosine}                                                     &                                                                                       & \textbf{0.0925}                                                 & 0.5901               & 0.4694                                                         & \textbf{0.7943}                                                        & \textbf{0.4510}                                                           & 0.0                                                              \\ \midrule
\textbf{R2 random}                                                     & \multirow{4}{*}{172}                                                                  & 0.1798                                                          & 0.7264               & 0.7153                                                         & 0.8391                                                                 & 0.5006                                                                    & 0.0                                                              \\
\textbf{R2 entropy}                                                    &                                                                                       & 0.2022                                                          & \textbf{0.7326}      & \textbf{0.7233}                                                & \textbf{0.8439}                                                        & 0.4877                                                                    & 0.1042                                                           \\
\textbf{R2 euclidean}                                                  &                                                                                       & \textbf{0.2311}                                                 & 0.7034               & 0.7063                                                         & 0.8362                                                                 & 0.5118                                                                    & \textbf{0.1447}                                                  \\
\textbf{R2 cosine}                                                     &                                                                                       & 0.2173                                                          & 0.7032               & 0.7063                                                         & 0.8390                                                                 & \textbf{0.5122}                                                           & 0.1215                                                           \\ \midrule
\textbf{R3 random}                                                     & \multirow{4}{*}{222}                                                                  & 0.2094                                                          & 0.7497               & 0.7065                                                         & 0.8501                                                                 & 0.5131                                                                    & 0.0994                                                           \\
\textbf{R3 entropy}                                                    &                                                                                       & 0.2407                                                          & 0.7500               & \textbf{0.7578}                                                & \textbf{0.8617}                                                        & 0.5224                                                                    & 0.2671                                                           \\
\textbf{R3 euclidean}                                                  &                                                                                       & 0.2609                                                          & 0.7497               & 0.7452                                                         & 0.8544                                                                 & \textbf{0.5452}                                                           & \textbf{0.3529}                                                  \\
\textbf{R3 cosine}                                                     &                                                                                       & \textbf{0.2616}                                                 & \textbf{0.7501}      & 0.7425                                                         & 0.8569                                                                 & 0.5449                                                                    & 0.3391                                                           \\ \midrule
\textbf{All data}                                                      & 440                                                                                   & 0.2523                                                          & 0.7749               & 0.7413                                                         & 0.8694                                                                 & 0.5386                                                                    & 0.1417                                                           \\ \bottomrule
\end{tabular}%
}
\end{table}

We further compare the performance of six critical target classes out of 25 classes for LC surgery in Table \ref{tab:my-table}, and the results of the full training set are in the last row.
The IoU in bold refers to the better strategy for corresponding class in each round. 
The table shows that different strategies have different influences depending on the class properties.
\begin{itemize}
  \item For the scissors tip, we observe a remarkable improvement when the model is trained under the active learning selected dataset. The surgeons suggest that scissors tip has a very similar texture and shape to the grasper tip while rarely exists in the dataset. For such underrepresented structures, active learning can select more frames with their presence since the model cannot perform well on such frames.
  \item For tiny structures like Johann grasper tip and cystic duct, we can see that feature distances perform better than entropy. It suggests the average entropy may down weight the significance of such structures even if the model is uncertain on these regions.
  \item For the well-defined and less morphologically different classes like diathermy hook tip and gallbladder, the result indicates that different informativeness measurement strategies influence less on the common and major classes.
\end{itemize}

Also, from the mean IoU and class-wise performance comparison, we can see that after the third active learning round, the model performance is approximate to which trained on the full training set. Note that, after the third round, the training set size is only 50\% of the full training set. 

\begin{figure}[!ht]
     \centering
     \begin{subfigure}[b]{0.45\linewidth}
         \centering
         \includegraphics[width=\textwidth]{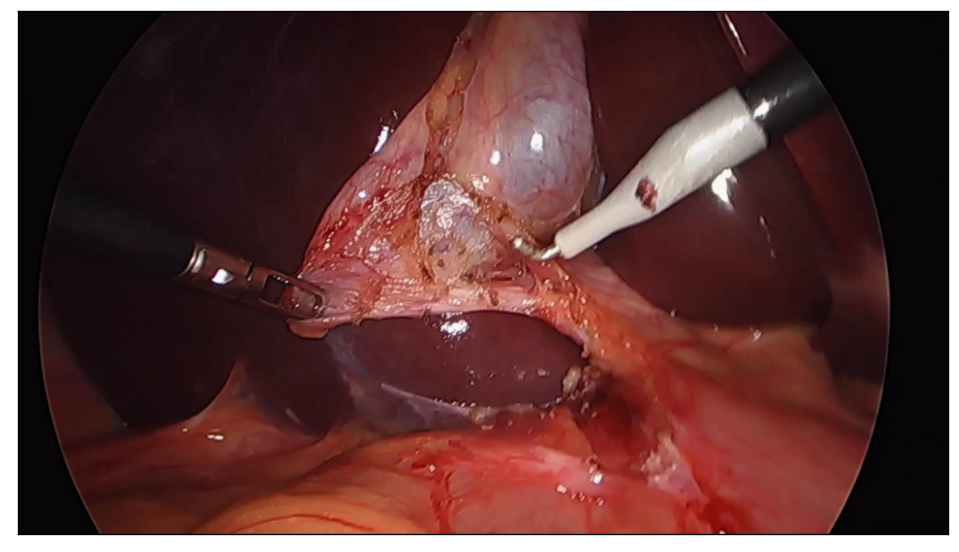}
         \caption{input frame}
         \label{fig:sf1}
     \end{subfigure}
     \hfill
     \begin{subfigure}[b]{0.45\linewidth}
         \centering
         \includegraphics[width=\textwidth]{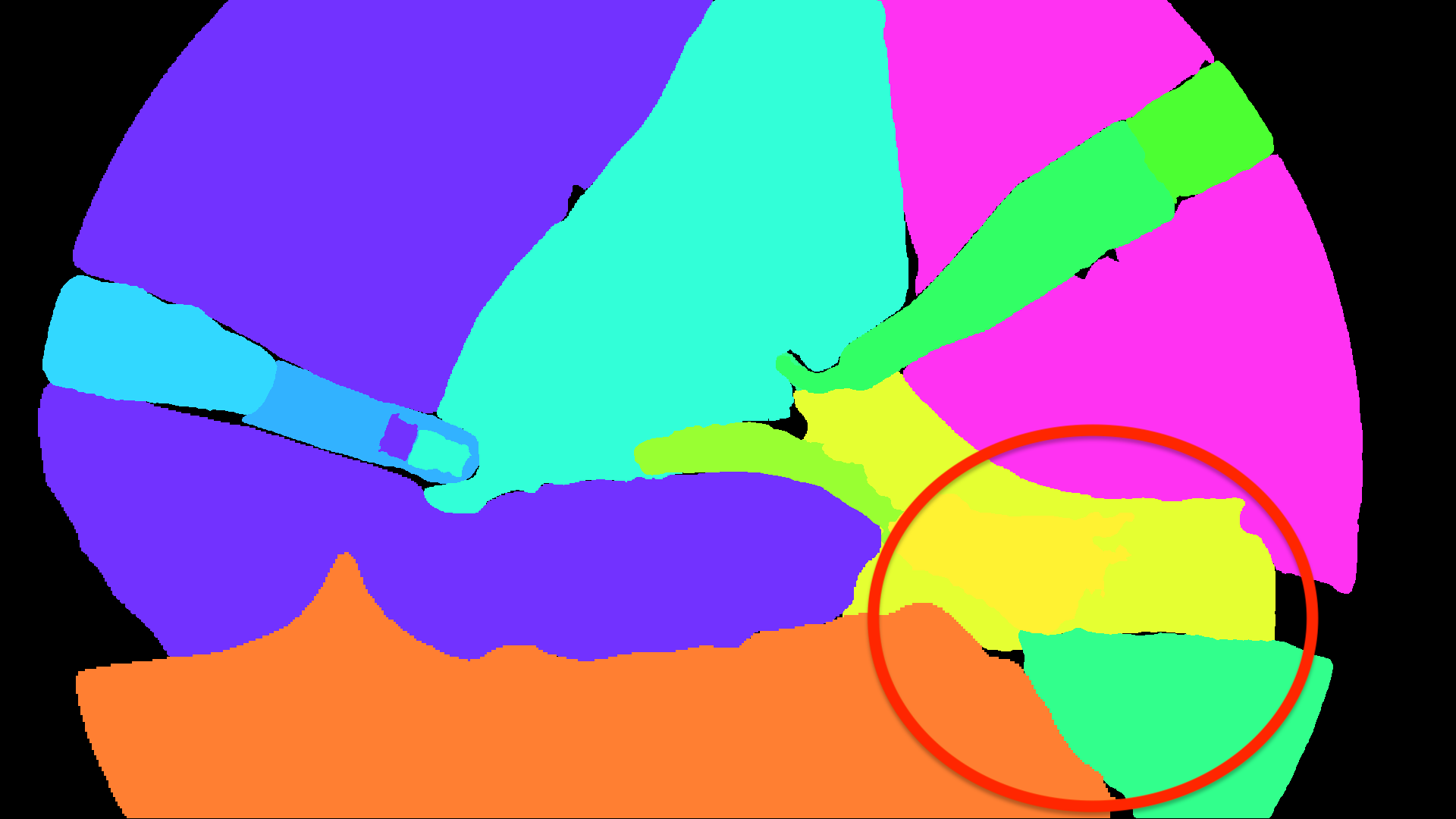}
         \caption{annotation}
         \label{fig:sf2}
     \end{subfigure}
     \hfill
     \begin{subfigure}[b]{0.45\linewidth}
         \centering
         \includegraphics[width=\textwidth]{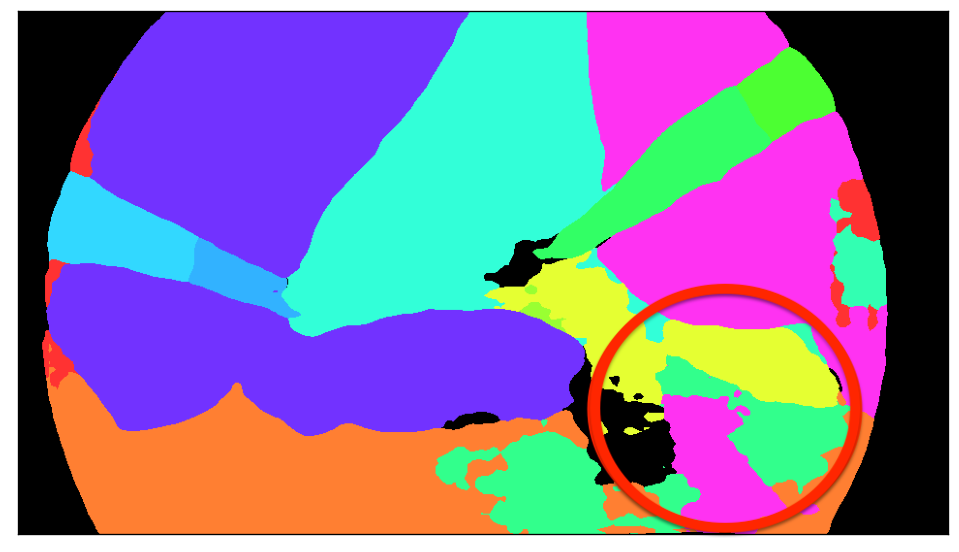}
         \caption{Round No.3 Random}
         \label{fig:sf3}
     \end{subfigure}
     \hfill
     \begin{subfigure}[b]{0.45\linewidth}
         \centering
         \includegraphics[width=\textwidth]{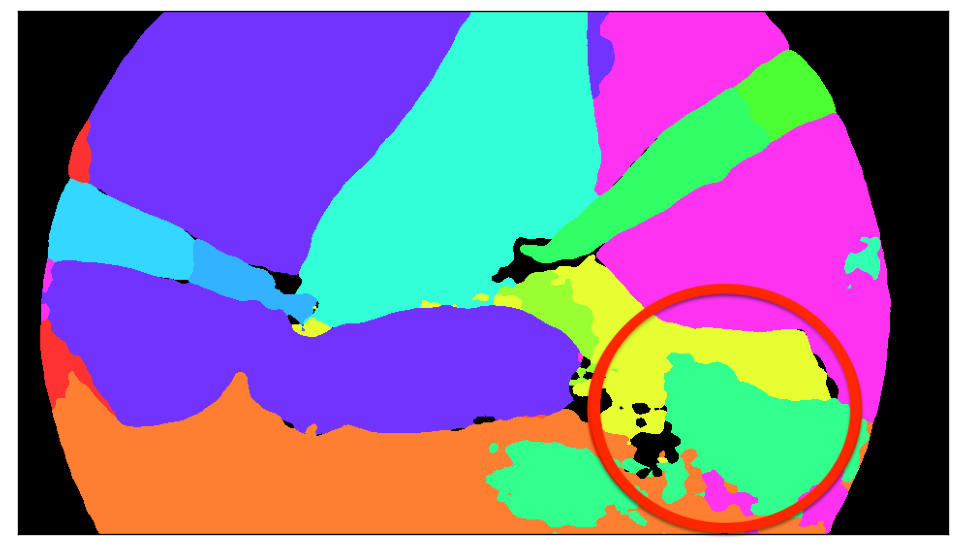}
         \caption{Round No.3 Entropy}
         \label{fig:sf4}
     \end{subfigure}
     \hfill
     \begin{subfigure}[b]{0.45\linewidth}
         \centering
         \includegraphics[width=\textwidth]{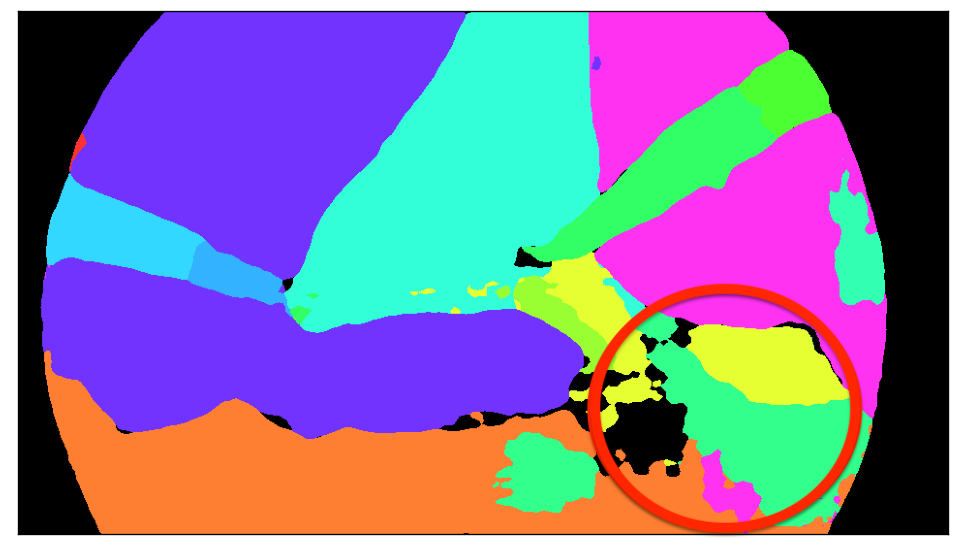}
         \caption{Round No.3 Cosine}
         \label{fig:sf5}
     \end{subfigure}
     \hfill
     \begin{subfigure}[b]{0.45\linewidth}
         \centering
         \includegraphics[width=\textwidth]{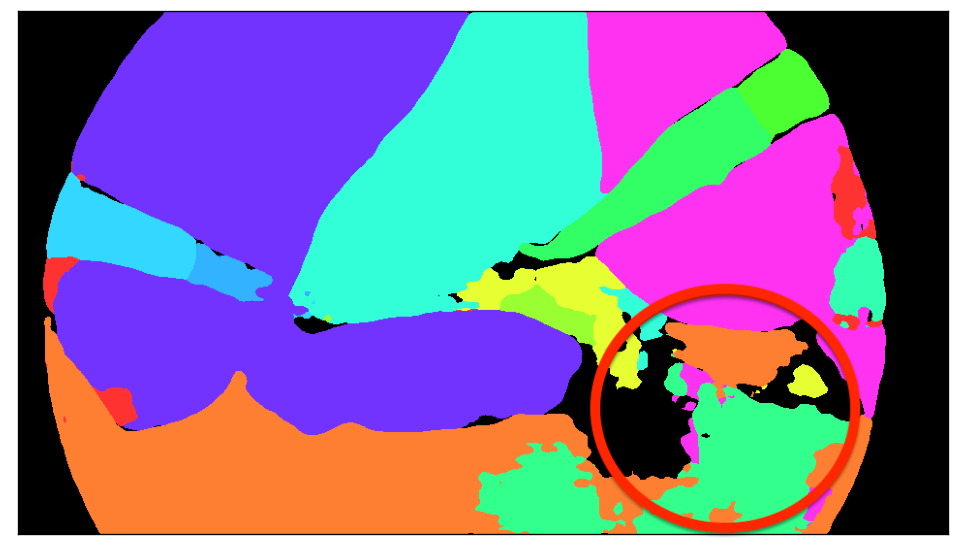}
         \caption{All data}
         \label{fig:sf6}
     \end{subfigure}
     \hfill
        \caption{Segmentation mask visualization }
        \label{fig:visualization}
\end{figure}

Figure. \ref{fig:visualization} shows the a sample test input prediction from models trained under different datasets after three selection rounds.
We demonstrated the input frame and its annotation in Figure. \ref{fig:sf1} and Figure. \ref{fig:sf2}, with the 4 prediction masks from random selection, entropy, and Cosine distance selected datasets, and ale the full training set in Figure. \ref{fig:sf3} to Figure. \ref{fig:sf6}.
We can notices that for area in the red circle, the model trained under Cosine feature distance selected dataset make the least error, followed by the entropy selection.
We also notices in Figure. \ref{fig:sf4} and Figure. \ref{fig:sf5} that the model trained under the informativeness selected datasets are better at predicting the well-represented regions like the liver (in pink); and the bowel and omentum (in orange) comparing to the model trained under all data.
It may imply that the informativeness measures can balance the dataset by both over-sampling the minor classes and under-sampling the abundant classes.

In this paper, the preliminary dataset is limited to only 5 video recordings. 
In the future, we will conduct expanded experiments on more intra- and inter-institution videos to verify our methods with more active learning iterations. 

Furthermore, this paper focuses on applying and accessing limited informativeness evaluation strategies separately. 
Siddiqui at el., proposed an uncertainty measurement based on the region entropy which shows promising performance in natural image semantic segmentation \cite{siddiqui2020viewal}.
In future studies, we will keep exploring different the distribution diversity and uncertainty measurement in the context of LC surgery segmentation task.
Novel informativeness measurement is also worth exploring specifically for surgical video dataset construction.

\section{Conclusion}
\label{sec:conclusion}
Laparoscopic cholecystectomy surgery is a suitable scenario for the deep learning application, where the video recordings provide a rich volume of data for model training with independent collection process. 
Due to the annotation difficulties, wise frame selection methods are demanded to efficiently construct high-quality, affordable datasets. 
Our paper shows active learning can choose the informative and challenging frames to train a good segmentation model. 
This approach allows less human intervention for threshold calibration and can significantly reduce the labeling cost. 
We believe our work can shed a light on the future surgical video dataset construction and therefore accelerate the interdisciplinary collaboration between deep learning and surgery.

\section*{Acknowledgment}
This research was undertaken using the LIEF HPC-GPGPU Facility hosted at the University of Melbourne. This Facility was established with the assistance of LIEF Grant LE170100200. All data are provided with ethics approval through St Vincent’s Hospital (ref HREC/67934/SVHM-2020-235987).  

\bibliographystyle{IEEEtran}
\bibliography{IEEEabrv,references}
\end{document}